  \documentclass[12pt]{article}

  \usepackage{graphicx}
  \usepackage{amssymb}
  \usepackage{textcomp}
  \usepackage{endnotes}
  \usepackage{color}
  \usepackage[colorlinks=true,urlcolor=blue]{hyperref}
  \usepackage{lineno}
  \usepackage{colortbl}

  \usepackage{amsmath}
  \usepackage{caption}
  \usepackage{comment}
  \usepackage{subcaption}
  \usepackage{longtable}
  \usepackage[font=sf]{caption}

  \graphicspath{{pictures/}}

  \usepackage{sectsty}
  \sectionfont{\fontsize{14}{18}\selectfont}

  \usepackage{titlesec}
  \titlespacing\section{0pt}{12pt plus 4pt minus 2pt}{0pt plus 2pt minus 2pt}

\begin{document}

  \let\footnote=\endnote
  \renewcommand{\notesname}{\bf Notes and References\vspace{-1ex}}

  \setlength{\fboxsep}{0pt}
  \setlength{\fboxrule}{1pt}

  
\begin{center} \LARGE Generating Synthetic Satellite Imagery \\[0.3ex] With Deep-Learning Text-to-Image Models \\[0.8ex] \large Technical Challenges and Implications for Monitoring and Verification

\end{center}

\begin{center} 

Tuong Vy Nguyen,$^*$
Alexander Glaser,$^{\dagger,\ddagger}$ and
Felix Biessmann$^{*,\ddagger}$

{\footnotesize \em $^*$Berliner Hochschule f\"ur Technik} \\
{\footnotesize \em $^\dagger$Program on Science and Global Security, Princeton University} \\
{\footnotesize \em $^\ddagger$Einstein Center Digital Future, Berlin} \\
\end{center}

{\bf Abstract.} Novel deep-learning (DL) architectures have reached a level where they can generate digital media, including photorealistic images, that are difficult to distinguish from real data. These technologies have already been used to generate training data for Machine Learning (ML) models, and large text-to-image models like DALL\raisebox{1pt}{$\cdot$}E 2, Imagen, and Stable Diffusion are achieving remarkable results in realistic high-resolution image generation. Given these developments, issues of data authentication in monitoring and verification deserve a careful and systematic analysis: How realistic are synthetic images? How easily can they be generated? How useful are they for ML researchers, and what is their potential for Open Science? In this work, we use novel DL models to explore how synthetic satellite images can be created using conditioning mechanisms. We investigate the challenges of synthetic satellite image generation and evaluate the results based on authenticity and state-of-the-art metrics. Furthermore, we investigate how synthetic data can alleviate the lack of data in the context of ML methods for remote-sensing. Finally we discuss implications of synthetic satellite imagery in the context of monitoring and verification.


\section{Introduction}

Novel technology for generating media content with automated and scalable machine learning (ML) technology has become a focus of attention in many application domains.
In a future where digital data from a variety of sources are abundant and widely available to non-governmental experts and independent analysts, and where virtually any type of digital media can be generated in ways that can make them effectively indistinguishable from real data, issues of data authentication in monitoring and verification deserve a careful and systematic analysis. 
While the impact of these developments on security aspects has been studied for text data in the context of misinformation in online social networks,\footnote{Ben Buchanan, Andrew Lohn, Micah Musser, Katerina Sedova, \href{https://cset.georgetown.edu/wp-content/uploads/CSET-Truth-Lies-and-Automation.pdf}{\em Truth, Lies, and Automation}, Center for Security and Emerging Technology, Georgetown University, Washington, DC, May 2021.} the challenges as well as the potential of synthesized data in the image domain have been underrepresented so far.

In this study, we assess the potential and risks associated with novel image generation methods such as in the context of monitoring and verification with a focus on satellite imagery. 

More concretely we first investigate the technological requirements to apply and adapt state of the art technology in image generation for the application scenario of nuclear verification with satellite imagery. In order to assess how easily the image generation can be controlled, we then generate a heterogeneous set of synthetic images and change a number of semantic dimensions in the image generation, such as seasonality, image context and time of day. We evaluate the quality of the generated images using established metrics and metrics that were specifically adapted to the task of remote sensing data.


\section{Datasets} \label{chap:data}

To compare the capabilities of the different approaches over a wider range of classes, two different datasets comprised of various objects, are leveraged.

\begin{figure}[bht]
\centering
\includegraphics[height=20mm]{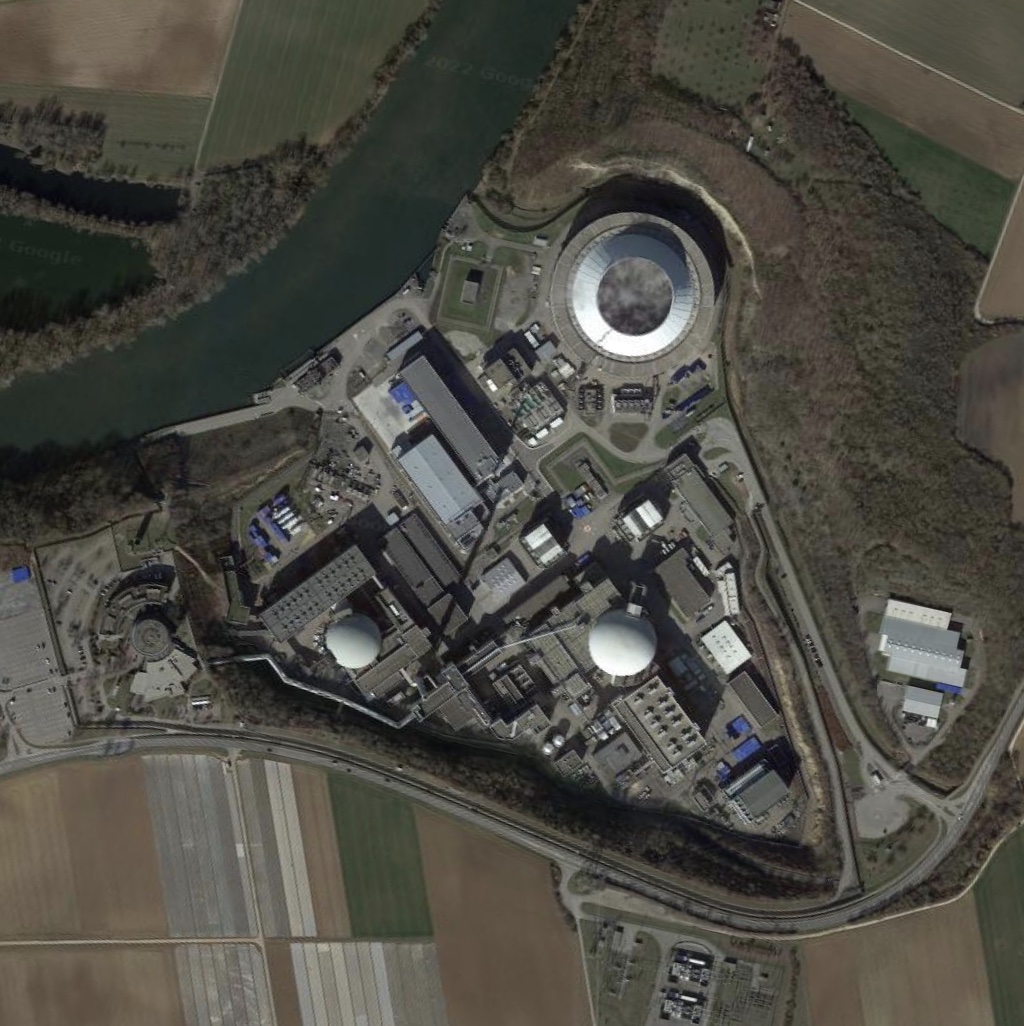}\hspace{2pt}%
\includegraphics[height=20mm]{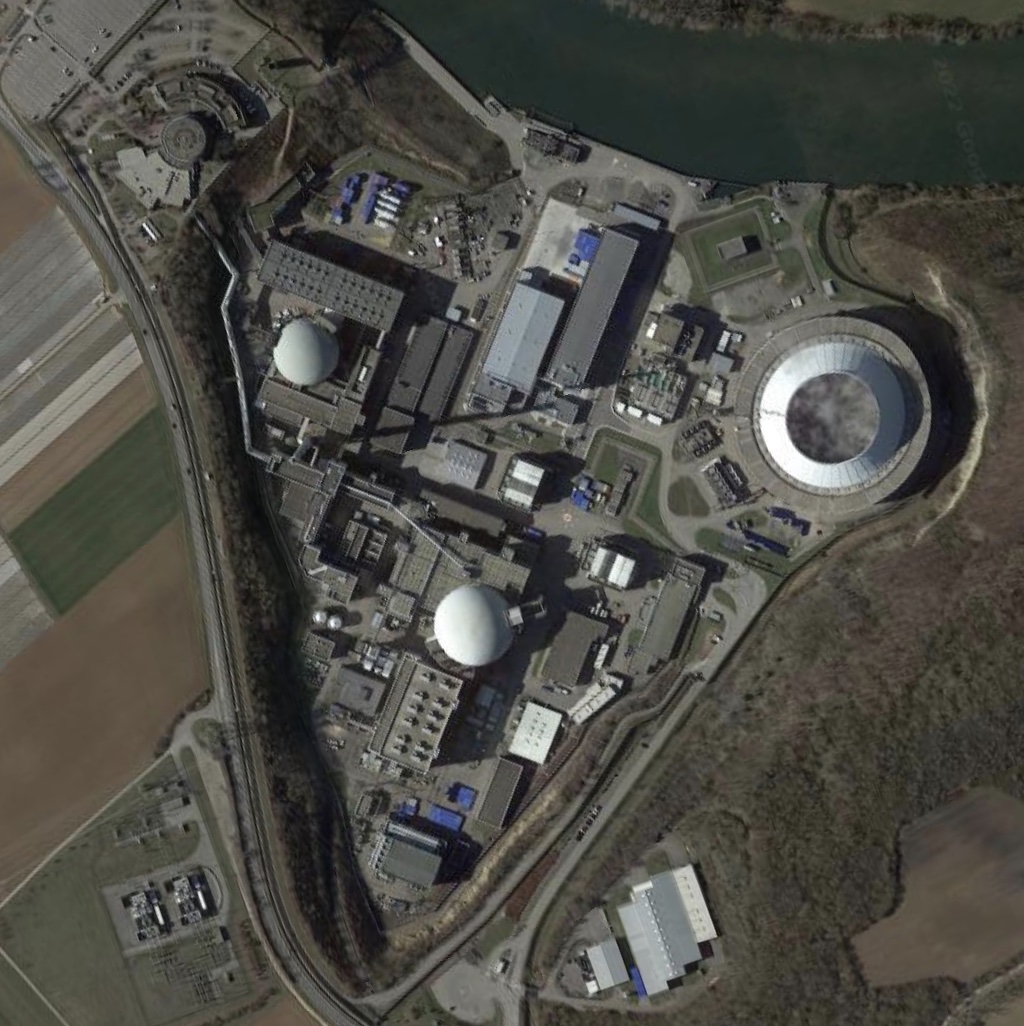}\hspace{2pt}%
\includegraphics[height=20mm]{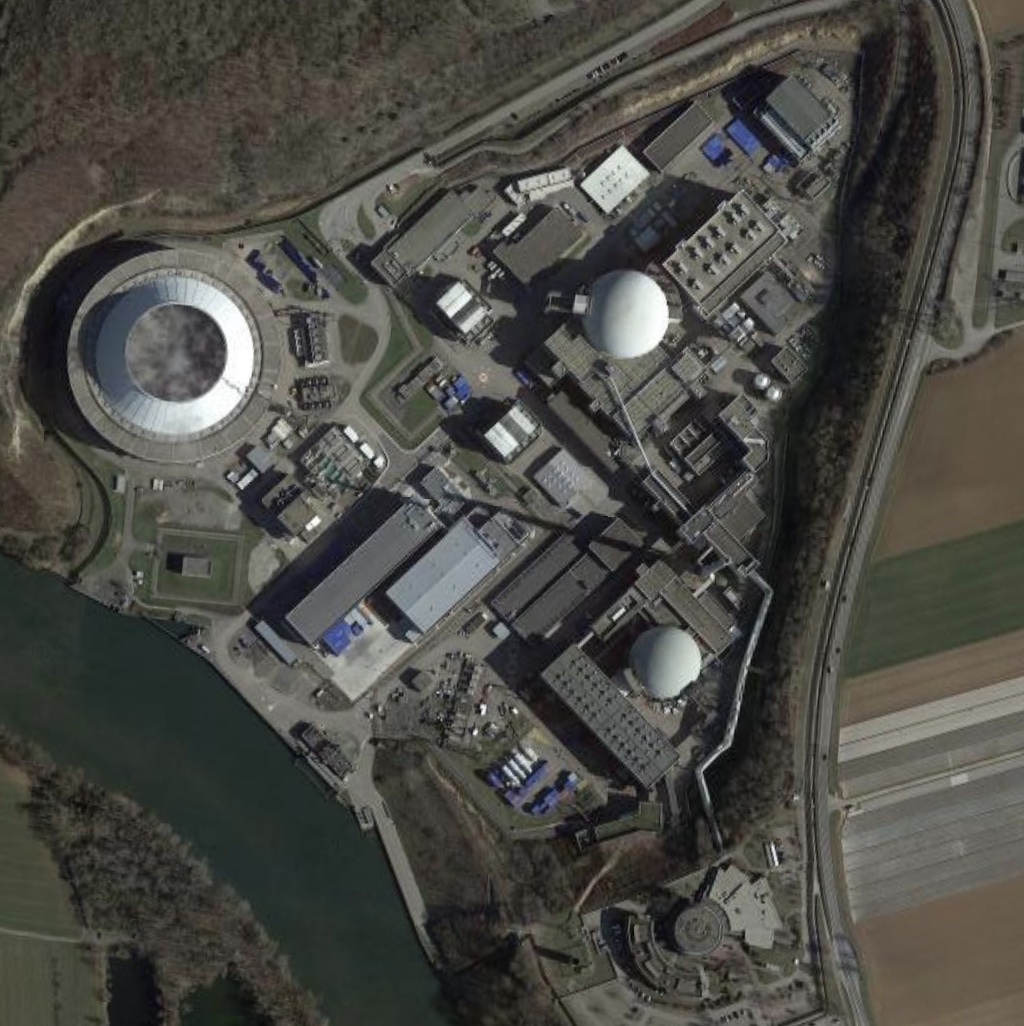}\hspace{15pt}%
\includegraphics[height=20mm]{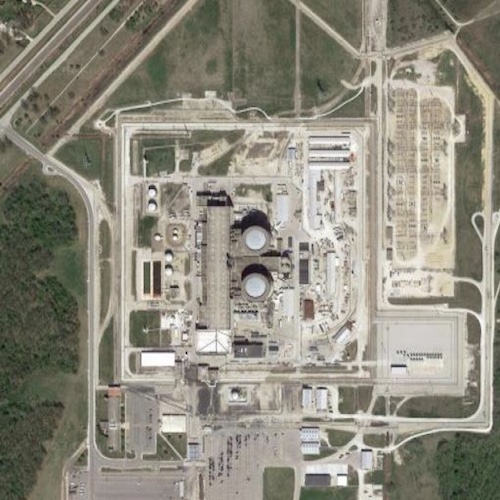}\hspace{2pt}%
\includegraphics[height=20mm]{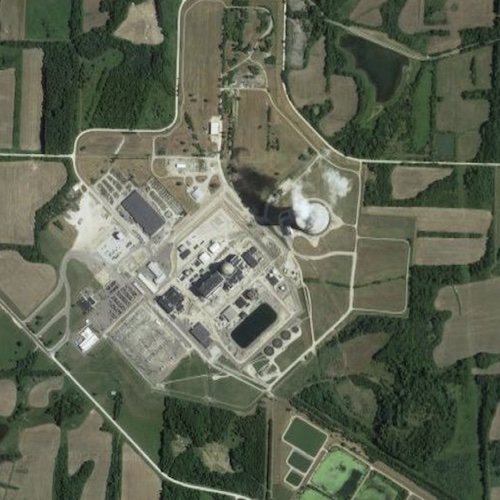}\hspace{2pt}%
\includegraphics[height=20mm]{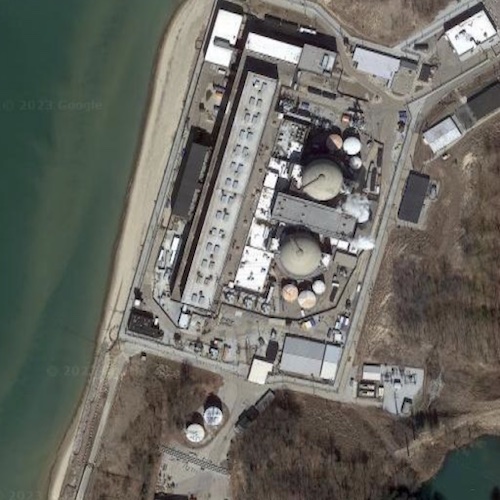}\hspace{2pt}%
\includegraphics[height=20mm]{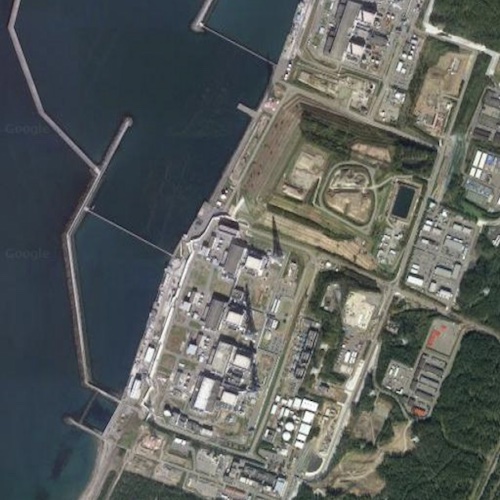}\\%
\includegraphics[height=20mm]{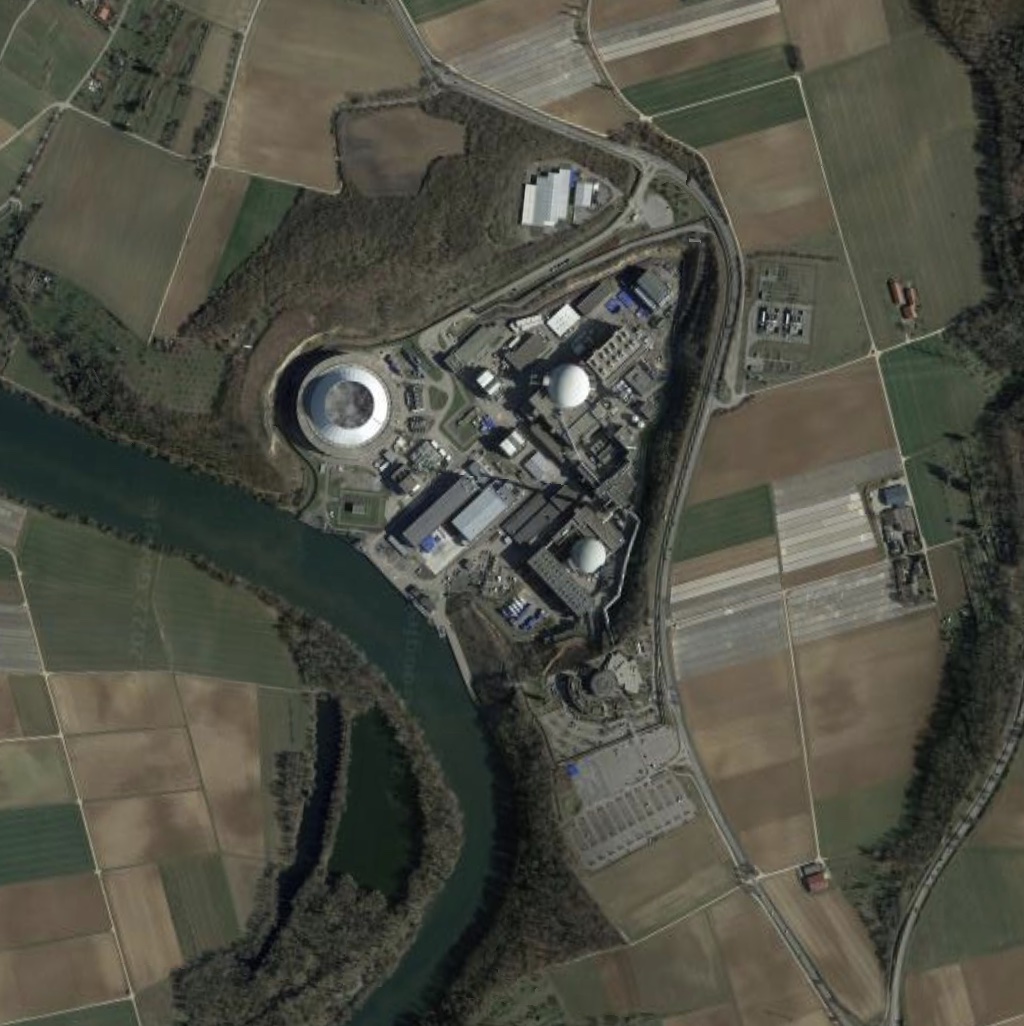}\hspace{2pt}%
\includegraphics[height=20mm]{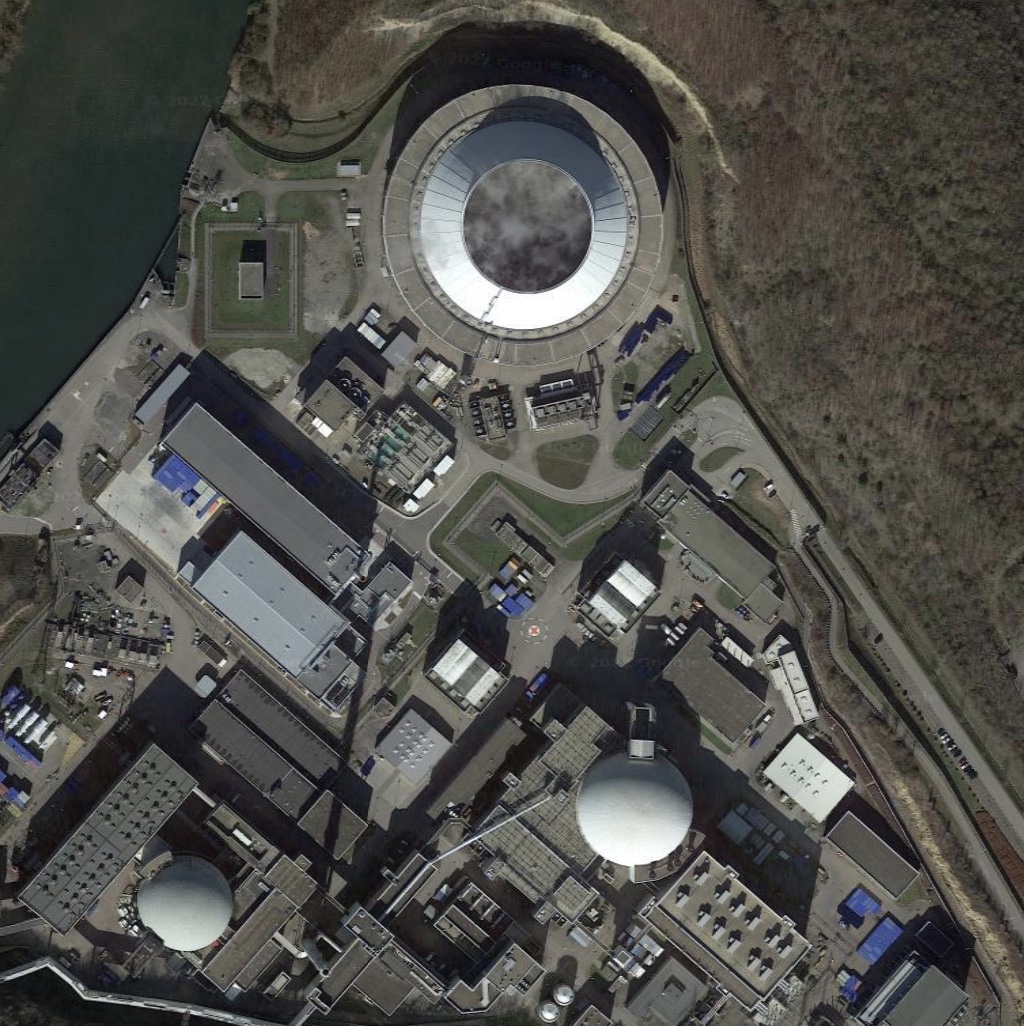}\hspace{2pt}%
\includegraphics[height=20mm]{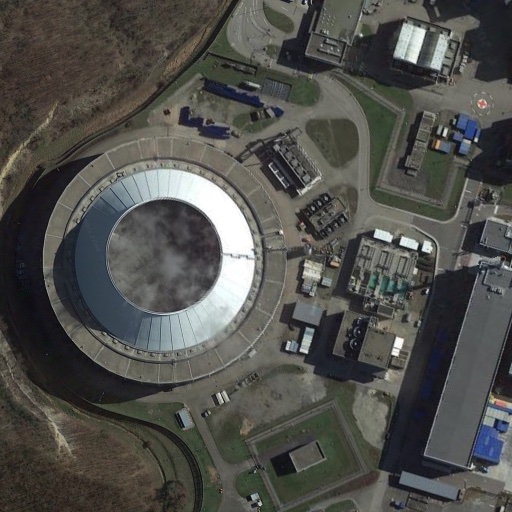}\hspace{15pt}%
\includegraphics[height=20mm]{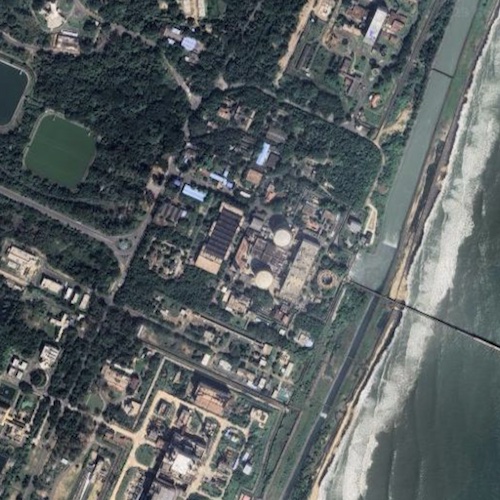}\hspace{2pt}%
\includegraphics[height=20mm]{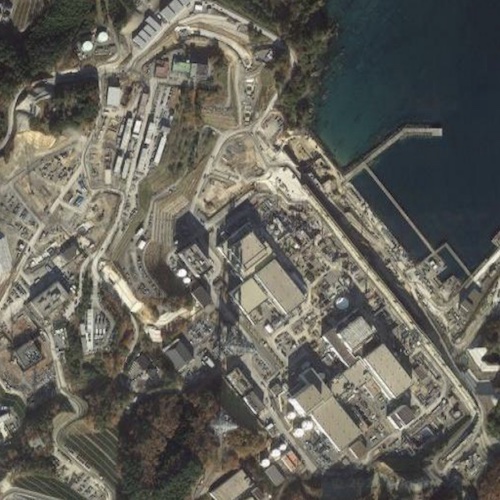}\hspace{2pt}%
\includegraphics[height=20mm]{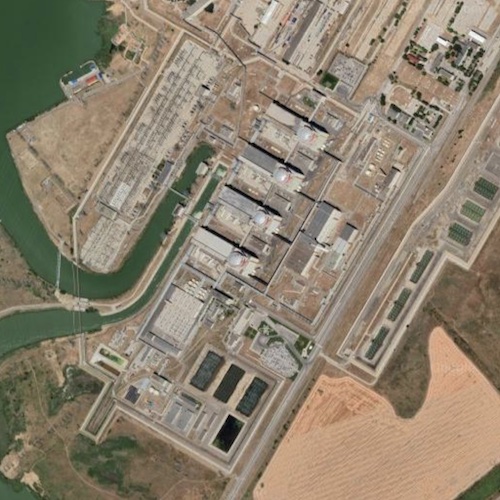}\hspace{2pt}%
\includegraphics[height=20mm]{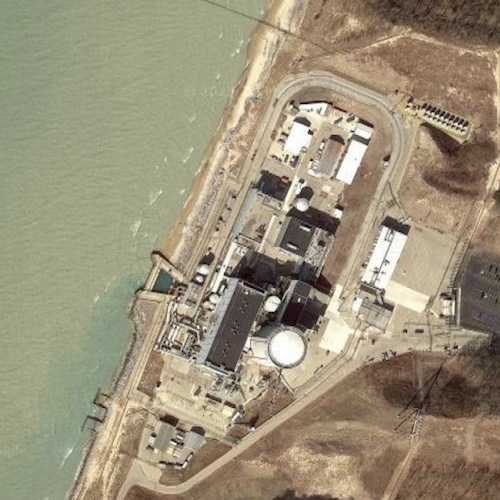}%
\caption{Shown on the left are six variations of the input image of the Neckarwestheim nuclear power plant; on the right are eight sample images  of nuclear power plants from different regions of the world. Overall, there are 202 input images in our dataset. {\it Source: Google Earth.}}
\label{fig:nuclear-power-plants}
\end{figure}

\textbf{\em Nuclear Facilities.} We use two different approaches to [fine-tune/train] the algorithms and generate synthetic imagery of nuclear power plants. First, we use different versions of a single input image of the Neckarwestheim nuclear power plant in Germany by varying the zoom factor and the rotation of the image (Figure~\ref{fig:nuclear-power-plants}, left). Second, we obtain training data from a large number of nuclear power plants using the Google Earth Engine (EE) in combination with web scraping tools.\footnote{For this imagery,  we used the ``Planet SkySat Public Ortho Imagery'' image collection collected in 2015 by Planet Labs' SkySat satellites.} Images of sites that were blurred or of low quality were removed, leaving 202 images of 185 unique nuclear power plants, which were then used for fine-tuning (Figure~\ref{fig:nuclear-power-plants}, right).


\textbf{\em UC Merced Land-use Dataset.} Another dataset used for our fine-tuning approaches is the UC Merced (UCM) dataset,\footnote{Yi Yang and Shawn Newsam, \href{http://weegee.vision.ucmerced.edu/datasets/landuse.html}{Bag-Of-Visual-Words and Spatial Extensions for Land-Use Classification}, {\em ACM SIGSPATIAL International Conference on Advances in Geographic Information Systems (ACM GIS),} 2010.} a benchmark land-use classification dataset from 2010, containing extracted images from the USGS National Map Urban Area Imagery collection of urban areas in the United States. The UCM images measure 256$\times$256 pixels and have a resolution of about 30~cm. The set contains 21 different classes with 100 images each for a total of 2100 images. 
Text captions for these images complement the dataset.\footnote{Bo Qu, Xuelong Li, Dacheng Tao, Xiaoqiang Lu, \href{https://ieeexplore.ieee.org/document/7546397}{Deep Semantic Understanding of High Resolution Remote Sensing Image}, {\em 2016 International Conference on Computer, Information and Telecommunication Systems (CITS),} 2016.} The splitting for training and testing was done according to the pre-defined split in the captions set, using 1680 images for fine-tuning and 210 each for validation and testing, respectively.


\section{Methods and Implementation}

Research on generative models for images has made dramatic progress over the past few years. For a long time, the state of the art in this novel domain have been Generative Adversarial Networks (GAN),\footnote{Ian J. Goodfellow et al., ``Generative Adversarial Nets,'' \href{https://arxiv.org/abs/1406.2661}{arXiv:1406.2661}, June 2014.} which have been explored thoroughly and applied in various domains. More recently, novel approaches to image generation based on Diffusion Models (DM) overcame some of the challenges associated with GAN training, resulting in a number of models that achieved higher image quality.\footnote{Jascha Sohl-Dickstein, Eric A. Weiss, Niru Maheswaranathan, Surya Ganguli, ``Deep Unsupervised Learning using Nonequilibrium Thermodynamics,'' \href{https://arxiv.org/abs/1503.03585}{arXiv:1503.03585}, March 2015; Jonathan Ho, Ajay Jain, Pieter Abbeel, ``Denoising Diffusion Probabilistic Models,'' \href{https://arxiv.org/abs/2006.11239}{arXiv:2006.11239}, June 2020.} Google's Imagen and OpenAI's DALL\raisebox{1pt}{$\cdot$}E 2 are two well-known examples.\footnote{Chitwan Saharia et al.,``Photorealistic Text-to-Image Diffusion Models with Deep Language Understanding,'' \href{https://arxiv.org/abs/2205.11487}{arXiv:2205.11487}, May 2022; Aditya Ramesh, Prafulla Dhariwal, Alex Nichol, Casey Chu, Mark Chen, ``Hierarchical Text-Conditional Image Generation with CLIP Latents,'' \href{https://arxiv.org/abs/2204.06125}{arXiv:2204.06125}, April 2022.} For our work, we have opted for Stable Diffusion as the underlying pre-trained text-to-image model for our purposes, as the source code and model weights have been made publicly available unlike with the other two models, and additional libraries and tools have been published as well, making the implementation fairly easy. Stable Diffusion is based on Latent Diffusion Models (LDM).\footnote{Robin Rombach, Andreas Blattmann, Dominik Lorenz, Patrick Esser, Björn Ommer, ``High-Resolution Image Synthesis with Latent Diffusion Models,'' \href{https://arxiv.org/abs/2112.10752}{arXiv:2112.10752}, December 2021.}

\begin{figure}
\includegraphics[width=\textwidth]{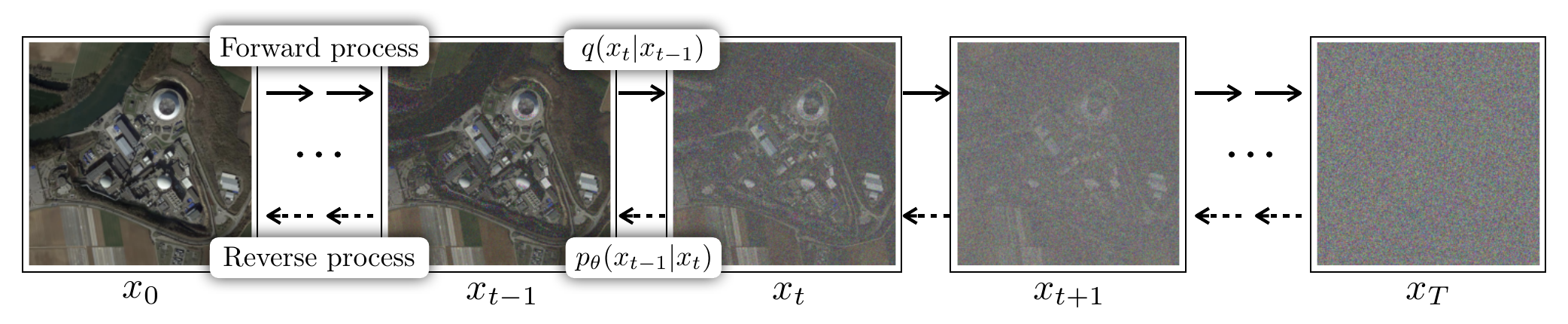}
\caption{Basic principle of Diffusion Models. During the first stage, the forward noising process, images are gradually perturbed by adding Gaussian noise. In the second stage, the reverse denoising process, a neural network is then learned to remove the noise iteratively to retrieve the original image.}
\label{fig:diffusion}
\end{figure}

Diffusion Models consist of two processes illustrated in Figure~\ref{fig:diffusion}. Moreover, LDMs make use of an Autoencoder architecture, transforming the images into a lower-dimensional latent space first before inputting this into the DM component. This allows for reduction of computational needs while still retaining image quality on similar level. For conditioning, a pre-trained text encoder is leveraged to transform the text prompts into embeddings and use them as conditioning inputs for the model. These and the perturbed latent representations are then used to train the denoising text-to-image U-Net component of Stable Diffusion. The obtained latents are then put through the Autoencoder's decoder to retrieve a sample back in pixel space.

To test the generalizability and the model's zero-shot capabilities, we use the unmodified pre-trained model as a baseline to examine how well the prior knowledge can be leveraged to generate satellite and aerial imagery. We then apply several fine-tuning approaches to further train the model on the datasets described above.
For the case of nuclear facilities, we then investigate what and how well conditionings can be applied to those newly learned concepts by adding keywords to the text prompts for variations regarding the location, seasonality and the time of day, for example, generating images of a nuclear power plant in the desert or in the winter. We also apply the fine-tuning methods on our aerial imagery dataset, UCM, which is comprised of more general known objects. 

For implementation, we use three different approaches: DreamBooth (DB),\footnote{Nataniel Ruiz, Yuanzhen Li, Varun Jampani, Yael Pritch, Michael Rubinstein, Kfir Aberman, ``DreamBooth: Fine Tuning Text-to-Image Diffusion Models for Subject-Driven Generation,'' \href{https://arxiv.org/abs/2208.12242}{arXiv:2208.12242}, August 2022.} Textual Inversion (TI),\footnote{Rinon Gal, Yuval Alaluf, Yuval Atzmon, Or Patashnik, Amit H. Bermano, Gal Chechik, Daniel Cohen-Or, ``An Image is Worth One Word: Personalizing Text-to-Image Generation using Textual Inversion,'' \href{https://arxiv.org/abs/2208.01618}{arXiv:2208.01618}, August 2022.} and Text-to-Image (Text2Img) fine-tuning.\footnote{\href{https://huggingface.co/docs/diffusers/training/text2image}{huggingface.co/docs/diffusers/training/text2image}. } Results are then compared to our baseline, vanilla Stable Diffusion (SDiff).\footnote{As opposed to fine-tuned models, vanilla means our pre-trained base model has not been further modified and is used off-the-shelf.} 
DreamBooth and Text2Img train the U-Net component of the model, while Textual Inversion trains in the text embedding space of the model's text encoder. For the nuclear facilities, we train on both the single nuclear plant in Neckarwestheim and on all facilities, separately. For UCM, we train a model for every of the single 21 land-use classes, using five randomly selected input images each. Furthermore, Huggingface's Diffusers Library is heavily used for the training and generative processes.


\section{Results}

\begin{figure}[!hbt]
\includegraphics[width=\textwidth]{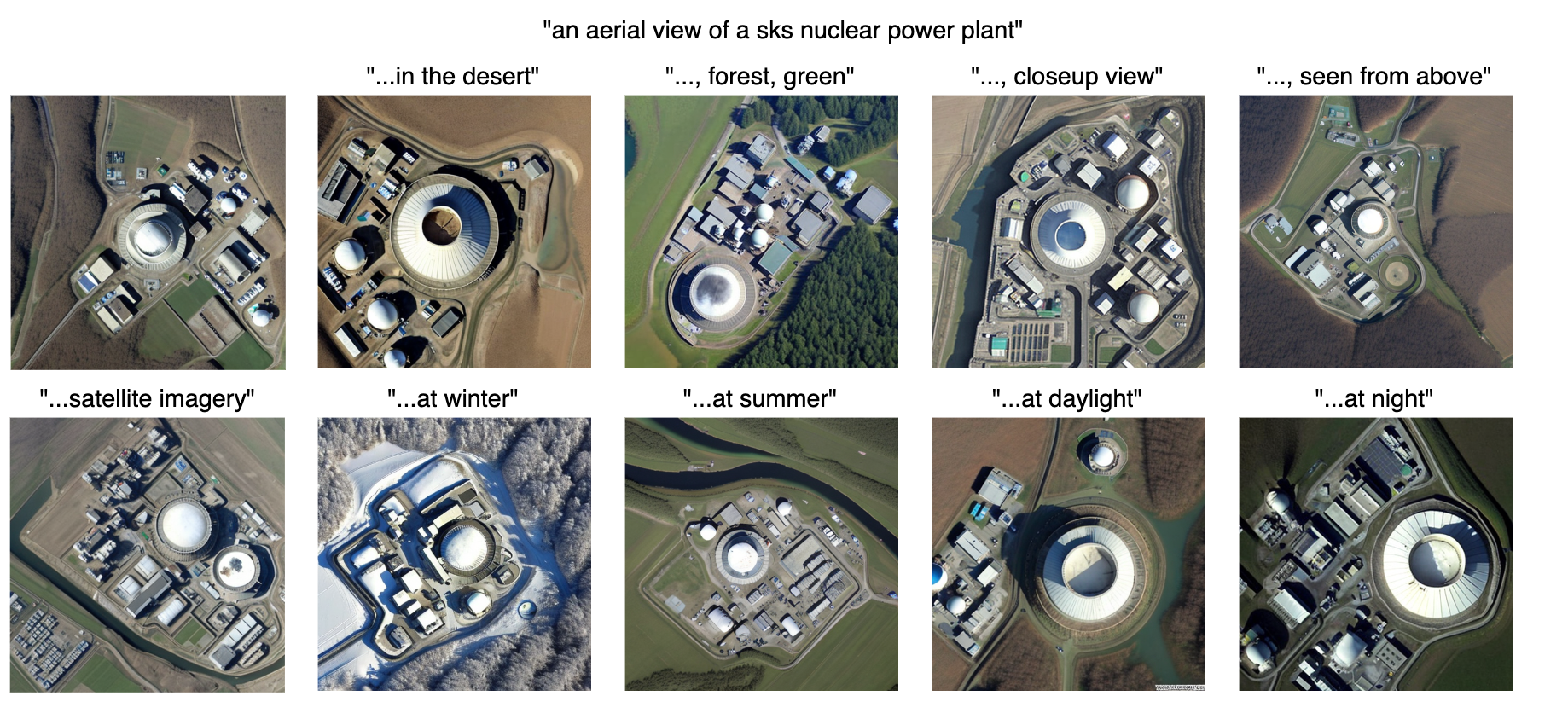} 
\caption{DreamBooth Neckarwestheim. Ten selected samples of synthetic images with their respective text prompts. Depicted are variations regarding seasonality, location, and time of day, all based on the original Neckarwestheim imagery from Figure~\ref{fig:nuclear-power-plants}.}
\label{fig:db-neckarwestheim}
\end{figure}

As part of this project, we have generated a large number of synthetic imagery. Ultimately, these images need to be characterized and assessed in order to draw conclusions about their usefulness for training purposes or other applications. Qualitatively, we have found that each method has their own strengths and weaknesses; these are discussed in greater detail in a forthcoming report.\footnote{Tuong Vy Nguyen, ``Machine Learning for Synthetic Satellite Images: Conditional Image Generation using a Vision-Language Model,'' Master’s Thesis, Berliner Hochschule für Technik, Berlin, May 15, 2023.} 
As an example, Figure~\ref{fig:db-neckarwestheim} shows some sample images produced with DreamBooth for the Neckarwestheim reactor sample; in general, DreamBooth is particularly strong at preserving image fidelity and reproducing distinct features. Furthermore, the fewer images are used as input, the better this seems to work for these subject-driven fine-tuning methods. This is plausible as with more input images, there are more specifics and differences to consider and preserving fine-grained details becomes more difficult when binding this to a single identifier or token.

For the UCM case, we evaluate the different methods based on a qualitative and quantitative assessment. Furthermore, a downstream task and user study are conducted. Looking at the generated samples for each applied method, we can observe that fine-tuning Stable Diffusion, our baseline and pre-trained text-to-image model, leads to imagery more suitable to the remote sensing domain. Furthermore, with the Text2Img model, the perspective from above, so our desired viewing angle, often seems to be retained as well, even for other objects or concepts, which are not present in the UCM dataset. 

\begin{figure}[!hbt]
\includegraphics[width=\textwidth]{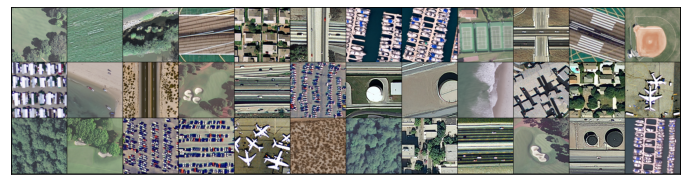}
\caption{UCM, Text2Img. Shown are 36 randomly selected samples of synthetic imagery. The data was generated with the Text2Img model, based on the text captions provided in the UCM dataset.}
\label{fig:xxx}
\end{figure}
  
Ultimately, however, a systematic assessment requires quantitive metrics for an evaluation of generative models and their synthesized data; the two main ones currently in use are the Inception Score (IS) and the Fr\'{e}chet Inception Distance (FID).\footnote{Tim Salimans, Ian Goodfellow, Wojciech Zaremba, Vicki Cheung, Alec Radford, Xi Chen, ``Improved Techniques for Training GANs,'' \href{https://arxiv.org/abs/1606.03498}{arXiv:1606.03498}, June 2016; Martin Heusel, Hubert Ramsauer, Thomas Unterthiner, Bernhard Nessler, Sepp Hochreiter, ``GANs Trained by a Two Time-Scale Update Rule Converge to a Local Nash Equilibrium,'' \href{https://arxiv.org/abs/1706.08500}{arXiv:1706.08500}, June 2017.}
The IS only assesses the synthetic images and evaluates image quality and diversity. The higher the score, the better. The FID compares the data distribution of real and fake images with each other and measures the dissimilarity, thus, the lower, the better. Both metrics rely on the feature space of a pre-trained Inception model, hence the name.
To adapt these metrics to our use cases, we exchange this underlying model with a pre-trained classifier model, which we further fine-tune on real UCM data, resulting in IS$^*$ and FID$^*$. The Fréchet Clip Distance (FCD) is another modification of the original FID, where the model is replaced by a pre-trained CLIP model.\footnote{Eyal Betzalel, Coby Penso, Aviv Navon, Ethan Fetaya, A Study on the Evaluation of Generative Models; betzalel-study-2022 and radford-clip-2021.}

\textbf{Nuclear facilities.} Due to lack of sufficient real data for comparison, only the IS and IS$^*$ are calculated for nuclear reactors considered in this study. Using different prompts where variations regarding the different dimensions seasonality, location, time of day and plain (no additional conditioning) are added to the text prompt. This is done for sample sizes $n=202$ and $n=6000$, respectively, to make it somewhat comparable to the real training images. The results are found in Table \ref{table:eval-results}. 



\begin{table}[!hbt]
\centering \sf
\scalebox{0.9}{
\begin{tabular}{ p{4.0cm} r r r r }
 \hline
 Model/Data & $\text{IS}_{202}$ $\uparrow$ & $\text{IS}^*_{202}$ $\uparrow$ & $\text{IS}_{6000}$ $\uparrow$ & $\text{IS}^*_{6000}$ $\uparrow$ \\
 \hline
 Real train images & 3.12 ± 0.44 & 3.80 ± 0.43 & --- & --- \\
 DB Neckar & 2.84 ± 0.38 & 2.13 ± 0.21 & 3.10 ± 0.08 & 2.26 ± 0.08 \\
 TI Neckar & \textbf{4.06 ± 0.42} & 3.52 ± 0.93 & \textbf{5.53 ± 0.11} & 4.13 ± 0.11 \\
 DB All & 2.30 ± 0.26 & 2.49 ± 0.28 & 2.60 ± 0.07 & 2.76 ± 0.05 \\
 TI All & 3.36 ± 0.36 & \textbf{4.26 ± 0.66} & 4.97 ± 0.13 & \textbf{5.39 ± 0.15} \\
 Van. SDiff & 2.99 ± 0.32 & 3.25 ± 0.59 & 3.75 ± 0.09 & 4.03 ± 0.12 \\
 \hline
\end{tabular}
}
\caption{Quantitative evaluation results. Scores for the nuclear power plant models have been calculated on the sample sizes 202 and 6000 each. $n=202$ was chosen so the scores are comparable to the IS of our real data (the images of nuclear power plants used for training). Larger values ($\uparrow$) are better.}
\label{table:eval-results}
\end{table}

While the results appear to suggest that the TI approach yields better scores than the pre-trained Vanilla Stable Diffusion model, it is difficult to draw conclusions from the IS and the modified IS$^*$ alone. Surprisingly, we find that the real satellite images used for training do not even achieve the best scores. This highlights the challenges associated with automated evaluation of generative methods, as well as the need to develop improved metrics. In addition, the two metrics do not reflect how well the generated images align with the text prompts or how accurately the semantic conditions and given variations can be displayed. Thus, a user study should be conducted to further investigate how the images are perceived by human evaluators.


\textbf{UC Merced Land-use Dataset.} For the quantitative assessment in the UCM case, we generate 420 samples for each method and compare their obtained quantitative results. It can be observed that for most of the metrics, the ranking is as follows: Text2Img outperforms the other models, followed by DB, then TI and, lastly, SDiff. This only deviates with the original IS, Recall and FID$^*$. Randomly selected samples of each approach are shown in the Appendix. These results indicate that if enough image data are available in a given domain to fine-tune the models used in the empirical evaluation of synthetic data as well as to calibrate the evaluation methods, the metrics we investigated are consistent in that they yield robust rankings of image generation methods. 

\begin{table}[!hbt]
\centering \sf
\scalebox{0.75}{
\begin{tabular}{ l r r r r r r r r  }
 \hline
 Model/Data & IS $\uparrow$ & IS$^*$ $\uparrow$ & FID $\downarrow$ & FID$^*$$\downarrow$ & FCD $\downarrow$ & KID $\downarrow$ & Precision $\uparrow$ & Recall $\uparrow$ \\
 \hline
 UCM Val+Test & 5.85±0.68 & 13.78±1.30 & --- & --- & --- & --- & --- & --- \\
 \hline
 Text2Img & 6.86±0.80 & \textbf{13.01±0.85} & \textbf{139.65} & 23.32 & \textbf{12.62} & \textbf{0.01±0.02} & \textbf{0.56} & 0.38 \\
 DB & 5.99±0.44 & 10.40±0.71 & 171.69 & 35.67 & 15.99 & 0.01±0.02 & 0.34 & 0.27 \\
 TI & 6.08±0.73 & 8.10±0.70 & 177.61 & \textbf{22.13} & 16.75 & 0.02±0.02 & 0.33 & 0.17 \\
 SDiff & \textbf{7.55±1.23} & 7.27±0.71 & 207.41 & 65.55 & 41.15 & 0.05±0.02 & 0.02 & \textbf{0.48} \\
 \hline
\end{tabular}
}
\caption{Quantitative evaluation results on UCM. The real UCM test and validation sets were used as one test set, resulting in $n=420$ real samples to compare against. Arrows indicate whether larger ($\uparrow$) or lower ($\downarrow$) values are better.}
\label{table:eval-results-ucm}
\end{table}

\textbf{UCM User Study.} For the user study, we used 210 images of each approach, resulting in 840 images. These were annotated by in total four users, with each image receiving one label. For four of the tasks the annotation was missing due to technical error, hence they were left out, leaving 836 annotated images to evaluate. The success rate is defined as the number of images that were classified as real for each method. The results of the user study are found in Table \ref{table:user-study-results}.

\begin{table}[!hbt]
\centering \sf
\scalebox{0.8}{
\begin{tabular}{ l | c c c c  }
 \hline
 Method & Real images & Text2Img & DB & TI \\
 \hline
 Predicted as ``Real'' & 158 & 119 & 115 & \phantom{0}69 \\
 Predicted as ``Fake''~~ & \phantom{0}52 & \phantom{0}89 & \phantom{0}94 & 140 \\
\hline
 Success Rate & \hspace{6mm}75\%\hspace{6mm} & \hspace{6mm}57\%\hspace{6mm} & \hspace{6mm}55\%\hspace{6mm} & \hspace{6mm}33\%\hspace{6mm} \\
\hline
\end{tabular}
}
\caption{User study results. Synthetic and real images were annotated as real or fake during an experiment with human evaluators. The success rate is shown for each approach. For example, out of the 210 real UCM images, 158 were classified as real, resulting in a success rate of 75\%; the other 52 images were labeled as fake, even though they were real.}
\label{table:user-study-results}
\end{table}

Overall, Text2img achieves the best results, followed by DB and TI. This ranking matches most of the applied quantitative evaluation metrics. Especially our adapted IS$^*$ aligns better with this than the original IS. Surprisingly, this is not the case for the adapted FID$^*$ though, where the original FID and FCD fit better. The specific reasons for this require further research, one assumption is that the similarity of real and synthetic images in the feature space captured by the neural networks doesn't necessarily say something about how real or authentic an image is actually perceived as and the used feature space might not be suitable either. Furthermore, the FID$^*$ scores still fluctuate for the used sample sizes, thus, don't seem to be as robust as the original FID, and the ranking might change with increasing sample size.

The results of the user study allow for a first comparison between the different evaluation metrics and how they correlate with human visual perception. Most importantly, the results demonstrate that all of the tested approaches can generate imagery that is, partly, able to fool the untrained human eye. Especially using DB or the Text2Img approach, where over 50\% of the synthetic imagery from both methods was labeled as real by human evaluators. 25\% of the real images were also classified as fake, meaning the generated images managed to mislead the users in that regard as well.


\section{Conclusion and Outlook}

Overall, it can be observed that, though not on par with real data, synthetic data can obtain evaluation scores of the same order of magnitude and we are able to, partly, generate realistic-looking images that can fool the human eye. Furthermore, fine-tuning the text-to-image model itself, so the U-Net component, seems to be the most effective space to further train, as this gives the best results (Text2Img and DB both do this). 

The development of suitable evaluation metrics for synthetic image quality is still an open field of research. Our results show that while in some cases, especially when not enough data is available to calibrate evaluation methods, existing evaluation metrics do not yield robust and reliable estimates of the synthetic image quality. At the same time our experiments demonstrate that if enough data is provided for calibrating quality metrics for synthetic imagery, existing metrics can work well: When adapted to the remote sensing domain, the metrics investigated yield robust rankings of image generation methods that aligns well with human perception. These findings underline the need for more research on quantitative evaluation metrics for synthetic data. We believe that this development requires a transdisciplinary effort including experts in deep learning and the specific domain in which data is generated.

 While the generation of artificial images is very interesting from a technical viewpoint, there are ethical and societal aspects to consider: Our results demonstrate that with open access to powerful models, technologies, and tools, low resource requirements and a feasible computing time, virtually anyone has the means to generate synthetic data at a larger scale with seemingly the touch of a button. Consequently, since image synthesis and also spoofing have become fairly easy to execute, the threat of malicious manipulations rises, with implementation becoming even accessible to non-experts and laymen. Using current methods, misinformation can quickly be spread through social media and various other outlets, and viewers are mislead by generated content. Recognizing deepfakes becomes increasingly difficult with the emerging generative technologies and non-technical solutions could also be considered, such as watermarking. 

Considering our domain, artificially generated satellite imagery, especially at a large scale, can alleviate the issue of inaccessible and unannotated data. However, this provides grounds for malicious purposes as synthetic satellite images, which contain real and fabricated features, are a notable threat regarding fake geography\footnote{Bo Zhao, Shaozeng Zhang, Chunxue Xu, Sun Yifan, Chengbin Deng, "Deep fake geography? When geospatial data encounter Artificial Intelligence", \href{https://doi.org/10.1080/15230406.2021.1910075}{10.1080/15230406.2021.1910075}, April 2021.}. We argue that the development of new generative approaches  should be accompanied by better methods to evaluate synthetic image quality and  detection of synthetic imagery, particularly in the remote sensing domain\footnote{Lydia Abady, Edoardo Daniele Cannas, Paolo Bestagini, Benedetta Tondi, Stefano Tubaro, and Mauro Barni, "An Overview on the Generation and Detection of Synthetic and Manipulated Satellite Images", \href{https://doi.org/10.48550/arXiv.2209.08984}{arXiv.2209.08984} September 2022.}.


Concluding, deepfakes can be a serious threat if containing compromising content and are circulated in public. With emerging technologies and tools, even non-experts have the means to generate large-scale synthetic datasets without proper safeguards set in place. The detection of fake data and a thorough investigation of the impact such images could have when widely distributed is an important aspect and should be further investigated in the future.


\begingroup 
\def\enotesize{\small}
\theendnotes
\endgroup

{\tiny ~ \hfill Revision 0}

  \end{document}